\documentclass[runningheads]{llncs}
\usepackage{graphicx}
\usepackage{amsmath}
\usepackage{amsfonts}
\usepackage{multirow}
\usepackage{algorithm}
\usepackage{cancel}
\usepackage{arydshln}
\usepackage[noend]{algpseudocode}
\usepackage[caption=false]{subfig}
\newcommand{\donotdisplay}[1]{}

\graphicspath{{Figures/}}

\usepackage{url}

\begin{document}
%
\title{Multivariate Powered Dirichlet-Hawkes Process}
\titlerunning{Multivariate Powered Dirichlet-Hawkes Process}
%
\author{Gaël Poux-Médard\inst{1}\orcidID{0000-0002-0103-8778} \and
Julien Velcin\inst{1}\orcidID{0000-0002-2262-045X} \and
Sabine Loudcher\inst{1}\orcidID{0000-0002-0494-0169}}
\authorrunning{G. Poux-Médard et al.}
%
\institute{$^1$ Université de Lyon, Lyon 2, ERIC UR 3083, 5 avenue Pierre Mendès France, F69676 Bron Cedex, France\\
\email{gael.poux-medard@univ-lyon2.fr}\\
\email{julien.velcin@univ-lyon2.fr}\\
\email{sabine.loudcher@univ-lyon2.fr}\\
}


%
\maketitle              
\begin{abstract}
The publication time of a document carries a relevant information about its semantic content. The Dirichlet-Hawkes process has been proposed to jointly model textual information and publication dynamics. This approach has been used with success in several recent works, and extended to tackle specific challenging problems --typically for short texts or entangled publication dynamics. However, the prior in its current form does not allow for complex publication dynamics. In particular, inferred topics are independent from each other --a publication about finance is assumed to have no influence on publications about politics, for instance.

In this work, we develop the Multivariate Powered Dirichlet-Hawkes Process (MPDHP), that alleviates this assumption. Publications about various topics can now influence each other. We detail and overcome the technical challenges that arise from considering interacting topics. We conduct a systematic evaluation of MPDHP on a range of synthetic datasets to define its application domain and limitations. Finally, we develop a use case of the MPDHP on Reddit data. At the end of this article, the interested reader will know how and when to use MPDHP, and when not to.

\end{abstract}
\section{Introduction}

\begin{figure}
    \centering
    \includegraphics[width=\columnwidth]{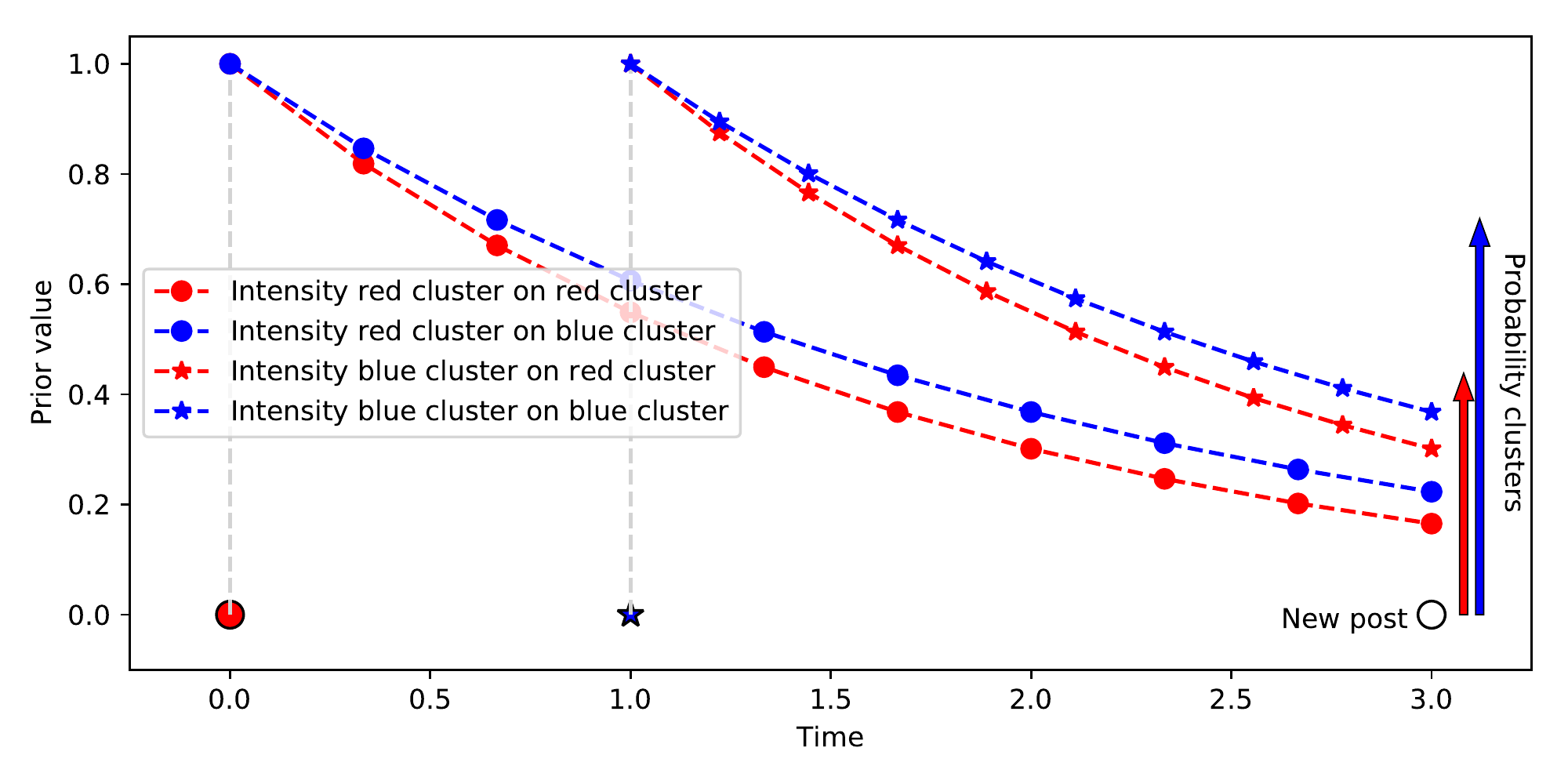}
    \caption{\textbf{Illustration of the Multivariate Powered Dirichlet-Hawkes Process prior} --- A new event appears at time $t=3$ from a cluster which is yet to be determined. The \textit{a priori} probability that this event belongs to a given cluster $c_{red}$ depends on the sum of the red intensity functions at time $t=3$. Similarly, the \textit{a priori} probability that this event belongs to a cluster $c_{blue}$ depends on the sum of the blue lines at time $t=3$. In previous models, this prior probability depends on each cluster self-stimulation only.}
    \label{fig:illustration-MPDHP}
\end{figure}

Understanding the data publication mechanisms on online platforms is of utmost importance in computer science. The amount of user-generated content that flows on social networks (e.g. Reddit) daily appeals for efficient and scalable approaches; they should provide us detailed insights within these mechanisms. A favoured approach to this problem is to cluster published documents according to their semantic content \cite{Blei2006DynamicTopicModel,Blei2003LDA,Yin2018ShortTextDHP}. 


In the specific case of data flowing on social networks, time also carries a valuable information about the underlying data flow generation process \cite{Du2015DHP,Valera2017HDHP,Tan2018IBHP}. Typically, we expect given publications to trigger ulterior publications within a short time range. This effect has been studied by considering spreading agents, who are individually influenced by contacts' publications \cite{GomezRodriguez2011NetRate,GomezRodriguez2013SurvivalAnalysis,Poux2023Houston}.

In previous works, the understanding of large data flows boils down to sorting data pieces (documents) into independent topics (clusters). However, it has been underlined on several occasions that online publication mechanisms are more complex than that. Typically, a correct description should involve clusters that interact with each other \cite{Myers2012CoC,Poux2021IMMSBM,Zarezade2017CorrelatedCascade}.
We illustrate the implications of this claim in Fig.~\ref{fig:illustration-MPDHP}. In most existing works that explicitly model both text and time, a given topic is assumed to only trigger observations from the same topic \cite{Du2015DHP,Poux2021PDHP} --the red cluster can only trigger observations from the red cluster. Instead, we must allow clusters to trigger publications in any other cluster.

Therefore, we extend a previous class of models (DHP \cite{Du2015DHP} and PDHP \cite{Poux2021PDHP}) to account for cluster interaction mechanisms. We show that technical challenges arise when considering topical interaction, and solve them. This results in the Multivariate Powered Dirichlet-Hawkes Process (MPDHP). We conduct systematic experiments to test the limits of MPDHP and define its application domain. In particular, we show that it performs well in cases when textual data is scarce and when the number of coexisting clusters is large. Finally, we investigate a real-world use case on a Reddit dataset.

\section{Background}
\label{SotA}

The original Dirichlet-Hawkes process (DHP) \cite{Du2015DHP} merges Dirichlet processes and Hawkes processes. It is used as a prior in Bayesian clustering along with a main model --typically a language model. The prior expresses the assumption that a new event from a given topic appears conditionally on the presence of older events from this same topic. The conditional probability is encoded into the intensity function of a Hawkes process. One such Hawkes process is associated to each topic. The temporal (Hawkes) intensity of a topic $c$ is written $\lambda_c (t \vert \mathcal{H}_c)$; it depends on the history of all previous events associated to topic $c$, written $\mathcal{H}_c$. If no Hawkes intensity manages to explain well enough the presence of a new observation happening further in time, the DHP \textit{a priori} guess is that the new observation belongs to a new cluster. DHP have first been used for automated summary generation \cite{Du2015DHP}. A list of textual documents appear in chronological order and are treated as such; the DHP infers clusters of documents that are based both on their textual content and their publication date, and studies their auto-stimulated publication dynamics. 
This process knew several developments, that essentially consider alternative Dirichlet-based priors combined with Hawkes processes --Hierarchical DHP \cite{Valera2017HDHP}, Indian Buffet Hawkes process \cite{Tan2018IBHP} and powered DHP \cite{Poux2021PDHP}.


However, in \cite{Poux2021PDHP,Yin2018ShortTextDHP}, the authors underline several limits of the standard Dirichlet-Hawkes processes and of the extensions mentioned earlier. For instance, DHP fails in cases where publications content carry few information: when textual content is short (e.g. tweets \cite{Yin2018ShortTextDHP}) or when vocabularies overlap significantly (e.g., topic-specific datasets). 
In \cite{Poux2021PDHP}, the problem is alleviated by considering a Powered Dirichlet process \cite{Poux2021PDP} instead of a standard Dirichlet process. This process is merged with a univariate Hawkes process to make the Powered Dirichlet-Hawkes process. The authors retrieve better results in challenging clustering situations (large temporal and textual overlaps).

However, none of these works allow clusters to interact with each other, despite clues pointing in that direction \cite{Myers2012CoC,Poux2021IMMSBM,Zarezade2017CorrelatedCascade}.
Indeed in \cite{Du2015DHP,Valera2017HDHP,Poux2021PDHP,Tan2018IBHP}, the considered Hawkes processes are univariate: a cluster can only be used to trigger events within itself. Exploring how clusters interact with each other would significantly extend our comprehension of the publication mechanisms at stake in various datasets --such as social media or scientific articles. Identifying which topics trigger the publication of other seemingly unrelated topics might be interesting in the study of fake news spreading. Understanding the dynamics at stake may help to surgically inhibit the spread of such topics using the right refutation. Another possible use case would be nudging users towards responsible behaviours regarding environment, health, tobacco, etc.

In this paper, we extend the (univariate) Powered Dirichlet-Hawkes Process to its multivariate version. There are several reasons why it has not been done in prior works: first of all, the adaptation to the multivariate case is not trivial and poses several technical challenges. As a \textbf{first contribution}, we detail the challenges that arise when developing the Multivariate Powered Dirichlet-Hawkes Process (MPDHP). We propose methods to overcome them while retaining a linear time complexity $\mathcal{O}(N)$. Doing so, we also relax the near-critical Hawkes process hypothesis made in \cite{Du2015DHP,Valera2017HDHP,Poux2021PDHP}. 
The second reason why the multivariate extension has not been developed in prior works, is that it greatly raises the number of parameters to estimate. The inference task might become harder, and the results irrelevant.
Our \textbf{second contribution} consists in a systematic evaluation of the MPDHP on a variety of synthetic situations. Our goal is to identify the limits of MPDHP regarding textual overlap, computation time, the amount of available data, the number of co-existing clusters, etc. We show that MPDHP is perfectly fit for solving a variety of challenging situations.
Finally, we illustrate the new insights on topical interaction obtained by running MPDHP on a real-world Reddit dataset.

\section{The Multivariate Powered Dirichlet-Hawkes process}
\subsection{Powered Dirichlet Process}
The Dirichlet process can be expressed using the Chinese Restaurant Process metaphor. Consider a restaurant with an infinity of empty tables. A first client enters the restaurant and sits to any of the empty tables with a probability proportional to $\alpha$ --the \textit{concentration parameter}. Another client then enters the restaurant, and sits either at one of the occupied tables with a probability linearly proportional to the number of clients already sat at the table, or to any of the empty tables with a probability proportional to $\alpha$. The process is then iterated for an infinite number of clients. The resulting clients distribution over the tables is equivalent to a draw from a Dirichlet distribution. The Powered Dirichlet process is intended as a generalisation of the Dirichlet process \cite{Poux2021PDP}.

The probability for a new client to sit at one of the occupied tables is now proportional to the number of clients at the power $r$. Let $C_i$ be the table chosen by the $i^{th}$ client and $\mathcal{H}$ the history of table allocation. Formally, the probability for the $i^{th}$ to choose a table reads:
\begin{equation}
    \label{eq-PDP}
    PDP (C_i = c \vert \alpha, r, \mathcal{H}) = 
    \begin{cases}
    \frac{N_c^r}{\alpha + \sum_c N_c^r} \text{ if c = 1, ..., K}\\
    \frac{\alpha}{\alpha + \sum_c N_c^r} \text{ if c = K+1}
    \end{cases}
\end{equation}
where $N_c$ is the number of people that already sat at table $c$, $K$ is the total number of tables, and $r$ a hyper-parameter. Note that when $r=1$ we recover the regular Dirichlet process, and when $r=0$ we recover the Uniform Process \cite{Wallach2010UnifP}.

\subsection{Multivariate Hawkes process}
A Hawkes process is a temporal point process where the appearance of new events is conditional on the realisation of previous events. It is fully characterised by an intensity function, noted $\lambda (t \vert \mathcal{H})$ that depends on the history of previous events. It is interpreted as the instantaneous probability of a new observation appearing: $\lambda(t)dt = P(e \in \left[ t,t+dt \right])$ with $e$ an event and $dt$ an infinitesimal time interval. For simplicity of notation, we omit the term $\mathcal{H}$ which is implicit anytime the intensity function $\lambda$ is mentioned.

As in the DHP \cite{Du2015DHP}, we define one Hawkes process for each cluster. However in DHP each of them is associated to a \textbf{univariate} Hawkes process, that depends only on the history of events comprised in this cluster. In our case, instead, we associate each cluster to a \textbf{multivariate} Hawkes process that depends on all the observations previous to the time being. Let $t_i^c$ be the time of realisation of the $i^{th}$ event belonging to cluster $c$. We write the intensity function for cluster $c$ at all times as:
\begin{equation}
    \label{eq-multiHawkes}
    \lambda_c (t) = \sum_{t_i^{c'} < t} \alpha_{c,c'} \cdot \kappa(t-t_i^{c'}) \ \ \ ; \ \ \  \kappa_l(\Delta t) = \frac{e^{-\frac{(\Delta t-\mu_l)^2}{2 \sigma_l^2}}}{\sqrt{2 \pi \sigma_l^2}}
\end{equation}
In Eq.~\ref{eq-multiHawkes}, $\alpha_{c,c'}$ is a vector of $L$ parameters to infer, and ${\kappa(t-t_i^{c'})}$ is a vector of $L$ temporal kernel functions depending only on the time difference between two events. In our case, we consider a Gaussian RBF kernel, that allows to model a range of different intensity functions.

The log-likelihood of a multivariate Hawkes process for all observations up to a time $T$ reads:
\begin{equation}
    \label{eq-lik-multiHawkes}
    \log \mathcal{L}(\alpha \vert \mathcal{H}) = \sum_c \int_{0}^T \lambda_c (t) dt + \sum_{t_i^{c}} \lambda_c (t_i^{c})
\end{equation}

\subsection{Multivariate Powered Dirichlet-Hawkes Process}
The Multivariate Powered Dirichlet-Hawkes Process (MPDHP) arises from the merging of the Powered Dirichlet Process (PDP) and the Multivariate Hawkes Process (MHP), described in the previous sections. As in \cite{Du2015DHP,Valera2017HDHP,Poux2021PDHP}, the counts in PDP are substituted with the intensity functions of a temporal point-process, here MHP. The \textit{a priori} probability that a new event is associated to a given cluster no longer depends on the population of this cluster, but on its temporal intensity at the time the new observation appears. This is illustrated in Fig.~\ref{fig:illustration-MPDHP}, where two events from two different clusters $c_{red}$ and $c_{blue}$ have already happened at times $t_0=0$ and $t_1=1$. A new event appears at time $t=3$. The \textit{a priori} probability that this event belongs to the cluster $c_{red}$ depends on the sum of the intensity functions of observations at $t_0$ and $t_1$ on cluster $c_{red}$ at time $t=3$ --sum of the red dotted lines. Similarly, the \textit{a priori} probability that this event belongs to the cluster $c_{blue}$ depends on the sum of the blue dotted lines at time $t=3$.

Let $t_i$ be the time at which the $i^{th}$ event appears. The resulting expression reads:
\begin{equation}
    \label{eq-MPDHP-prior}
    P(C_i = c\vert t_i, r, \lambda_0, \mathcal{H}) = 
    \begin{cases}
    \frac{\lambda_c^r(t_i)}{\lambda_0 + \sum_{c'} \lambda_{c'}^r(t_i)} \text{ if c$\leq$K}\\
    \frac{\lambda_0}{\lambda_0 + \sum_{c'} \lambda_{c'}^r(t_i)} \text{ if c=K+1}
    \end{cases}
\end{equation}
In Eq.~\ref{eq-MPDHP-prior}, $\lambda_c$ in defined as in Eq.~\ref{eq-multiHawkes}, and the parameter $\lambda_0$ is the equivalent of the concentration parameter described in Eq.~\ref{eq-PDP}. Taking back the illustration in Fig.~\ref{fig:illustration-MPDHP}, this parameter corresponds to a time-independent intensity function. It has a chance to get chosen typically when the other intensity functions are below it (meaning they do not manage to explain the dynamic aspect of a new event). In this case, a new topic is opened, and gets associated to its own intensity function.

\subsection{Language model}
Similarly to what has been done in \cite{Du2015DHP,Poux2021PDHP}, the MPDHP must be associated to a Bayesian model given it is a prior on sequential data. Since we study applications on textual data, we choose to side the MPDHP prior with the same Dirichlet-Multinomial language model as in previous publications \cite{Du2015DHP,Poux2021PDHP}. According to this model, the likelihood of the $i^{th}$ document belonging to cluster $c$ reads:
\begin{equation}
    \label{eq-MPDHP-text}
    \mathcal{L}(C_i=c \vert N_{<i,c}, n_i, \theta_0) =\frac{\Gamma(N_c+\theta_0)}{\Gamma(N_c+n_i+\theta_0)} \prod_v \frac{\Gamma(N_{c,v} + n_{i,v} + \theta_{0,v})}{\Gamma(N_{c,v}+\theta_0)}
\end{equation}
where $N_c$ is the total number of words in cluster $c$ from observations previous to $i$, $n_i$ is the total number of words in document $i$, $N_{c,v}$ the count of word $v$ in cluster $c$, $n_{i,v}$ the count of word $v$ in document $i$ and $\theta_0 = \sum_v \theta_{0,v}$. Note that for any empty cluster, the likelihood is computed using $N_{c_{empty}} = 0$ for every empty cluster $c_{empty}$.

\section{Implementation}
\subsection{Base algorithm}
\subsubsection{SMC algorithm}
We use a Sequential Monte Carlo (SMC) algorithm for the optimisation. The base algorithm is the same as in \cite{Du2015DHP,Valera2017HDHP,Poux2021PDHP} -- a graphical representation of the SMC algorithm is provided in \cite{Poux2021PDHP}-Fig.1 and as Supplementary Material.
The goal of the SMC algorithm is to jointly infer textual documents' clusters and the dynamics associated with them. It runs as follows. First, the algorithm computes each cluster's posterior probability for a new observation by multiplying the temporal prior on cluster allocation (see Eq.~\ref{eq-MPDHP-prior}, illustrated Fig.~\ref{fig:illustration-MPDHP}) with the textual likelihood (see Eq.~\ref{eq-MPDHP-text}). It results in an array of $K+1$ probabilities, where $K$ is the number of non-empty clusters. A cluster label is then sampled from this probability vector. 
If the empty $(K+1)^{th}$ cluster is chosen, the new observation is added to this cluster, and its dynamics are randomly initialised (i.e. a $(K+1)^{th}$ row and a $(K+1)^{th}$ column are added to the parameters matrix $\alpha$). If a non-empty cluster is chosen, its dynamics are updated by maximising the new likelihood Eq.~\ref{eq-lik-multiHawkes}. The process then goes on to the next observation.

This routine is repeated $N_{part}$ times in parallel. Each parallel run is referred to as a \textit{particle}. Each particle keeps track of a series of cluster allocation hypotheses. After an observation has been treated, we compute the particles likelihood given their respective cluster allocations hypotheses. Particles that have a likelihood relative to the other particles' one below a given threshold $\omega_{thres}$ are discarded and replaced by a more plausible existing particle.

\subsubsection{Sampling the temporal parameters}
The parameters $\alpha$ are inferred using a sampling procedure. A number $N_{sample}$ of precomputed vectors is drawn from a Dirichlet distribution with probability $P(\alpha \vert \alpha_0)$, with $\alpha_0$ a concentration parameter. As the SMC algorithm runs, within each existing cluster, each of these candidate vectors is associated to a likelihood computed from Eq.~\ref{eq-lik-multiHawkes}, noted $P(\mathcal{H} \vert \alpha)$, where $\mathcal{H}$ represents the data. The sampling procedure returns the average of each of the $N_{sample}$ precomputed $\alpha$, weighted by the posterior distribution associated to them ${P(\alpha \vert \mathcal{H}) \propto P(\mathcal{H} \vert \alpha)P(\alpha \vert \alpha_0)}$. The so-returned matrix is guaranteed to be a good statistical approximation of the optimal matrix, provided the number of sample matrices $N_{sample}$ is large enough.

\subsubsection{Limits}
This algorithm described here works well for the univariate case, but fails for the multivariate case. In particular, updating the multivariate intensity function of each cluster requires knowing the number of already existing clusters, which vary over time. Therefore, we cannot precompute the sample matrices in advance --they must be updated as the algorithm runs to account for the right number of non-empty clusters. Moreover, the number of parameters to estimate evolves linearly with the number of active clusters $K$, instead of remaining constant as in DHP and variants \cite{Du2015DHP,Poux2021PDHP}. Because the number of parameters is not constant anymore, their candidate values cannot be sampled from a Dirichlet distribution anymore. In the following, we review these challenges and present our solutions to overcome them. We manage to preserve a constant time complexity for each observation.

\subsection{Optimisation challenges}
\subsubsection{Temporal horizon}
A first problem that has been answered in \cite{Du2015DHP} is that, for each new observation, the algorithm has to run through the whole history of events to compute the DHP prior. However, carefully choosing the kernel vector $\kappa (\cdot)$ described in Eq.~\ref{eq-multiHawkes} allows to perform this step in constant time. If the chosen kernel vanishes as time goes, it happens a point where old events have a near-zero chance to have any influence on new observations, according to our model. These events can be discarded from the computation for new events.

\subsubsection{Updating the triggering kernels}
In the univariate case \cite{Du2015DHP,Valera2017HDHP,Poux2021PDHP}, the coefficients $\alpha_{c} \in \mathbb{R}^L$ are sampled from a collection of existing sample vectors computed at the beginning of the algorithm (where $L$ is the size of the kernel vector). However, we must now infer a matrix instead. We recall that matrix $\alpha_c$ represents the weights given to the temporal kernel vector of every cluster influence on $c$ --see Eq.~\ref{eq-multiHawkes}. The likelihood Eq.~\ref{eq-lik-multiHawkes} can be updated incrementally for each sample matrix. A given cluster $c$ has a likelihood value associated to each of those $N_{sample}$ sample matrices, which represents how fit one sample matrix is to explain one cluster's dynamics. The final value of the parameters matrix is sampled simply by averaging the samples matrices weighted by their likelihood for a given cluster times the prior probability of these vectors being drawn in the first place.

Such sampling was possible in the univariate case, where each sample matrix was in fact a vector of fixed length. In our case, because Hawkes processes are multivariate, each entry $\alpha_{c} \in \mathbb{R}^{K \times L}$ is now a matrix. Moreover, the number of existing clusters $K$ increases over with time, and can grow very large. Each time a new cluster is added to the computation, a row is appended to the $\alpha_c$ matrix --it accounts for the influence of this new cluster regarding $c$.

However, some older events can be discarded from the computation. When an event is older than $3\sigma$ with respect to the longest range entry of the RBF kernel, it can be safely discarded. Clusters whose last observation has been discarded thus have a near-zero chance to get sampled once again. These clusters' contribution to the likelihood Eq.~\ref{eq-lik-multiHawkes} will not change anymore. Therefore, they do not have a role in the computation of the parameters matrix $\alpha_c$. The row corresponding to each of these clusters in the parameters matrix can then be discarded in every sample matrix. 
Put differently, the last sampled value for their influence on $c$ will remain unchanged for the remaining of the algorithm. 
The dimension of $\alpha_c$ only depends on the number of \textit{active} clusters, whose intensity function has not faded to zero. For a given dataset, the number of active clusters typically fluctuates around a constant value, making one iteration running in constant time $\mathcal{O}(1)$.

\subsubsection{A beta prior on parameters}
Another problem inherent to the multivariate modelling is the prior assumption on sample vectors. In \cite{Du2015DHP,Poux2021PDHP}, each sample vector is sampled from a Dirichlet distribution. This choice is to infer Hawkes processes that are nearly-unstable: the spectral radius of the temporal kernel function $\lambda_{c}(t)$ is close to 1. However in our case, such assumption is not possible because the size of each sample matrix can vary as the number of active clusters evolve. Drawing one Dirichlet vector for each entry $\alpha_{c,c'}$ would force the spectral radius of $\alpha_c$ to equal $K=\vert \alpha_c \vert$, which transcribes a highly-unstable Hawkes process. Our solution is to consider the parameters as completely independent from each other. Each entry of the matrix $\alpha$ is drawn from an independent $\beta$ distribution of parameter $\beta_0$. In this way, we make no assumption on the spectral radius of the Hawkes process, and samples rows/columns corresponding to new clusters can be generated one after the other.

\section{Experiments}
\subsection{Setup}
We design a series of experiments to determine the use cases of the Multivariate Powered Dirichlet-Hawkes Process\footnote{Data and implementations are available at \url{https://github.com/GaelPouxMedard/MPDHP}}. 
We list the parameters we consider in our experiments. When a parameter does not explicitly vary, it takes a default value given between parenthesis. These parameters are: the textual overlap (0) and the temporal overlap (0) discussed further in the text,
the temporal concentration parameter $\lambda_0$ (0.01), the strength of temporal dependence $r$ (1), the number of synthetically generated clusters $K$ (2), the number of words associated to each document $n_{words}$ (20), the number of particles $N_{part}$ (10) and the number of sample matrices used for sampling $\alpha$, noted $N_{sample}$ (2~000). For the detail of these parameters, please refer to Eq.~\ref{eq-MPDHP-prior}. The interplay between the parameters that are not part of MPDHP ($N_{sample}$ and $N_{part}$) is studied in Supplementary Material.

\begin{figure}[h]
    \centering
    \includegraphics[width=0.95\textwidth]{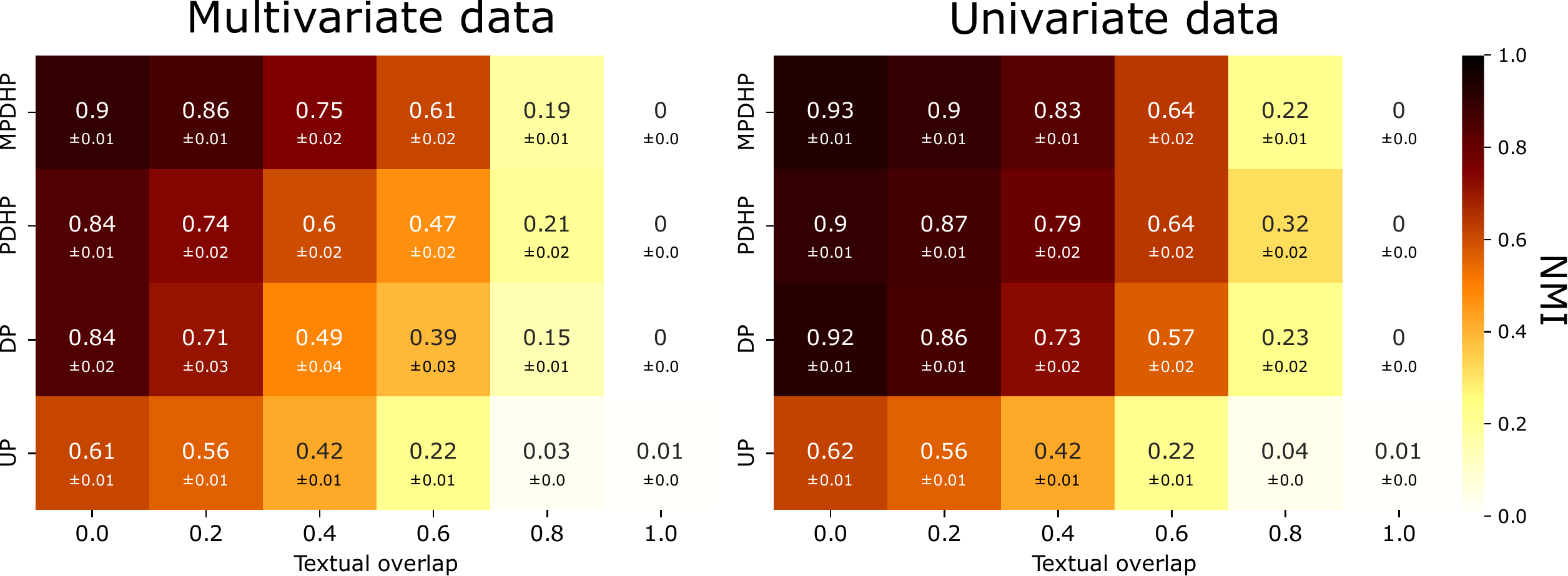}
    \caption{\textbf{Numerical results on synthetic data} --- MPDHP consistently outperforms other baselines designed for the univariate case on both univariate and multivariate data. The standard error has been computed using 100 independent runs.}
    \label{fig:mainres}
\end{figure}
Note that the overlap $o(f_1,...,f_N)$ between $N$ functions is defined as the sum over each function $f_i$ of its intersecting area with the largest of the $N-1$ other functions, divided by the sum of each function's total area \cite{Poux2021PDHP}. This value is bounded between 0 (perfectly separated functions) and 1 (identical functions):
\begin{align}
    \label{eq:overlap}
    o(f_1,...,f_N) = \ \sum_i \int_{\mathbb{R}} min(f_i(x), max(\{f_j(x)\}_{j \neq i})) dx \notag
\end{align}
For each combination of parameters considered, we generate 10 different datasets. In all datasets, we consider a fixed size vocabulary $V=1000$ for each cluster. All datasets are made of 5~000 observations. Observations for each cluster $c$ are generated using a RBF temporal kernel $\kappa (t)$ weighted by a parameter matrix $\alpha_c$. We set $\kappa (t) = \left[ \mathcal{G}(3;0.5) ; \mathcal{G}(7;0.5) ; \mathcal{G}(11;0.5) \right]$ where $\mathcal{G}(\mu;\sigma)$ is a Gaussian of mean $\mu$ and standard deviation $\sigma$.
We note $L=3$ the number of entries of $\kappa$. The inferred entries of $\alpha$ determine the amplitude (weight) of each kernel entry.

The generation process is as follows. First, we draw a random matrix $\alpha \in \mathbb{R}^{K \times L}$ and normalise it so that its spectral radius equals 1 --near unstable Hawkes process. We repeat this process until we obtain the wanted temporal overlap. Then, we simulate the multivariate Hawkes process using the triggering kernels $\alpha \cdot \kappa(t)$, where $\kappa (t)$ is the RBF kernel as defined earlier. Given the Hawkes process is multivariate, each event is associated to its class it has been generated from among $K$ possible classes. For each event, we draw $n_{words}$ words from a vocabulary of size $V$. The vocabularies are drawn from a multinomial distribution and shifted over this distribution so that they overlap to a given extent.

\subsection{Baselines}
We evaluate our clustering results in terms of Normalised Mutual Information score (NMI). This metric is standard when evaluating non-parametric clustering models. It compares two cluster partitions (i.e. the inferred and the ground truth ones); it is bounded between 0 (each true cluster is represented to the same extent in each of the inferred ones) and 1 (each inferred partition comprises 100\% of one true cluster). The standard error is computed on 100 runs. We compare our approach to 3 closely related baselines. \textbf{Powered Dirichlet-Hawkes process (PDHP)} \cite{Poux2021PDHP}: in this model, clusters can only self replicate. It means that the intensity function of a cluster $c$ Eq.~\ref{eq-multiHawkes} only considers past events that happened in the same cluster $c$. $r$ is set to 1. \textbf{Dirichlet process (DP)}: this prior is standard in clustering problems. It corresponds to a special case of Eq.~\ref{eq-PDP} where $r=1$. The prior probability for an observation to belong to a cluster depends on its population. \textbf{Uniform process (UP)} \cite{Wallach2010UnifP}: this prior corresponds to a special case of Eq.~\ref{eq-PDP} where $r=0$. It assumes that the prior probability for an observation to belong to a cluster does not depend on any information about this cluster (neither population nor dynamics).

\donotdisplay{
\begin{itemize}
    \item \textbf{Powered Dirichlet-Hawkes process (PDHP)} \cite{Poux2021PDHP} -- In this model, clusters can only self replicate. It means that the intensity function of a cluster $c$ Eq.~\ref{eq-multiHawkes} only considers past events that happened in the same cluster $c$. $r$ is set to 1.
    \item \textbf{Dirichlet process (DP)} -- This prior is standard in clustering problems. It corresponds to a special case of Eq.~\ref{eq-PDP} where $r=1$. The prior probability for an observation to belong to a cluster depends on its population.
    \item \textbf{Uniform process (UP)} \cite{Wallach2010UnifP} -- This prior corresponds to a special case of Eq.~\ref{eq-PDP} where $r=0$. It assumes that the prior probability for an observation to belong to a cluster does not depend on any information about this cluster (neither population nor dynamics).
\end{itemize}
}

\subsection{Results on synthetic data}
\subsubsection{MPDHP outperforms state-of-the-art} 
In Fig.~\ref{fig:mainres}, we plot our results for datasets that have been generated using a Multivariate Hawkes process (clusters have an influence on each other) and a Univariate Hawkes process (clusters can only influence themselves). We compare MPDHP to our baselines for various values of textual overlap --we provide a similar study that considers various temporal overlaps as Supplementary Material. 
\textbf{MPDHP} systematically outperforms the baselines on multivariate data --when clusters interact with each other. Considering that clusters interacts with each other improves our description of the datasets. \textbf{MPDHP} performs at least as good as PDHP on univariate data --when clusters can only self-stimulate. The complexity of MPDHP does not make it unfit to simpler tasks. \textbf{PDHP} performs better than MPDHP when the textual overlap is large (textual overlap of 0.8) due to its reduced complexity. Increasing the number of observations would fix this. 
\donotdisplay{
\begin{itemize}
    \item \textbf{MPDHP} systematically outperforms the baselines on multivariate data --when clusters interact with each other. Considering that clusters interacts with each other improves our description of the datasets. 
    \item \textbf{MPDHP} performs at least as good as PDHP on univariate data --when clusters can only self-stimulate. The complexity of MPDHP does not make it unfit to simpler tasks.
    \item \textbf{PDP} performs well for small textual overlaps, but fails when the textual overlap increases. Since only the textual information is considered by the PDP, this highlights the importance of also considering the temporal information.
    \item \textbf{PDHP} performs better than MPDHP when the textual overlap is large (textual overlap of 0.8) due to its reduced complexity. Increasing the number of observations would fix this.
    \item \textbf{All models} fail for full textual overlap; clusters cannot be inferred from temporal information only.
\end{itemize}
}

\subsubsection{Highly interacting processes} 
\begin{figure}
    \centering
    \includegraphics[width=\columnwidth]{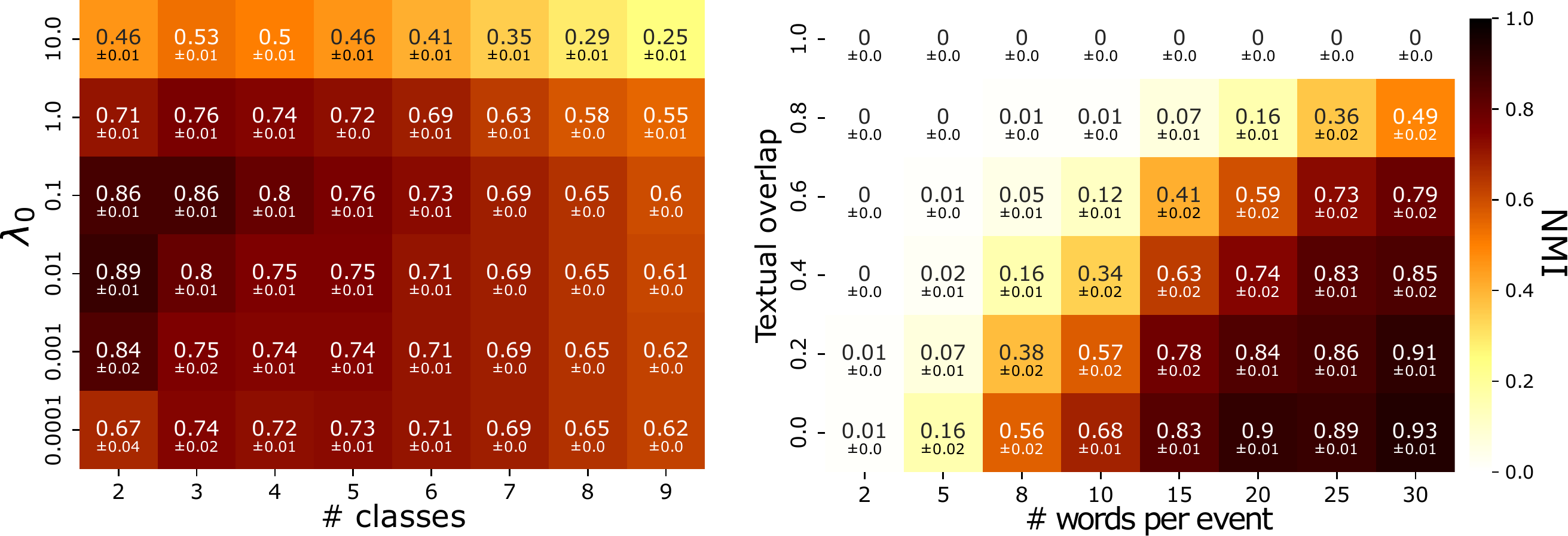}
    \caption{\textbf{MPDHP can handle a large number of coexisting clusters and scarce textual information} --- MPDHP yields good results when a large number of clusters coexist simultaneously (left) and when texts are short or little informative (right). It is also robust against variations of $\lambda_0$ over 5 order of magnitude.}
    \label{fig:XP2}
\end{figure}

We test when a large number of clusters coexist simultaneously. The rate at which new clusters get opened is mainly controlled by the $\lambda_0$ hyperparameter (see Eq.~\ref{eq-MPDHP-prior}), which we vary to see whether MPDHP is robust against it. In Fig.~\ref{fig:XP2} (left), we plot the performances of MPDHP according to these two parameters. We draw two conclusions: MPDHP can handle a large number of coexisting clusters and still correctly identify to which one each document belongs, and MPDHP is robust against large variations of $\lambda_0$. In this case, results are similar for $\lambda_0$ varying over 5 orders of magnitude. It means MPDHP does not have to be fine-tuned according to the number of expected clusters in cases where this number is not known in advance.
An extended discussion on the choice of the parameter $\lambda_0$ is provided as Supplementary Material.

\subsubsection{Handling scarce textual information} 
\donotdisplay{
\begin{figure}
    \centering
    \includegraphics[width=\columnwidth]{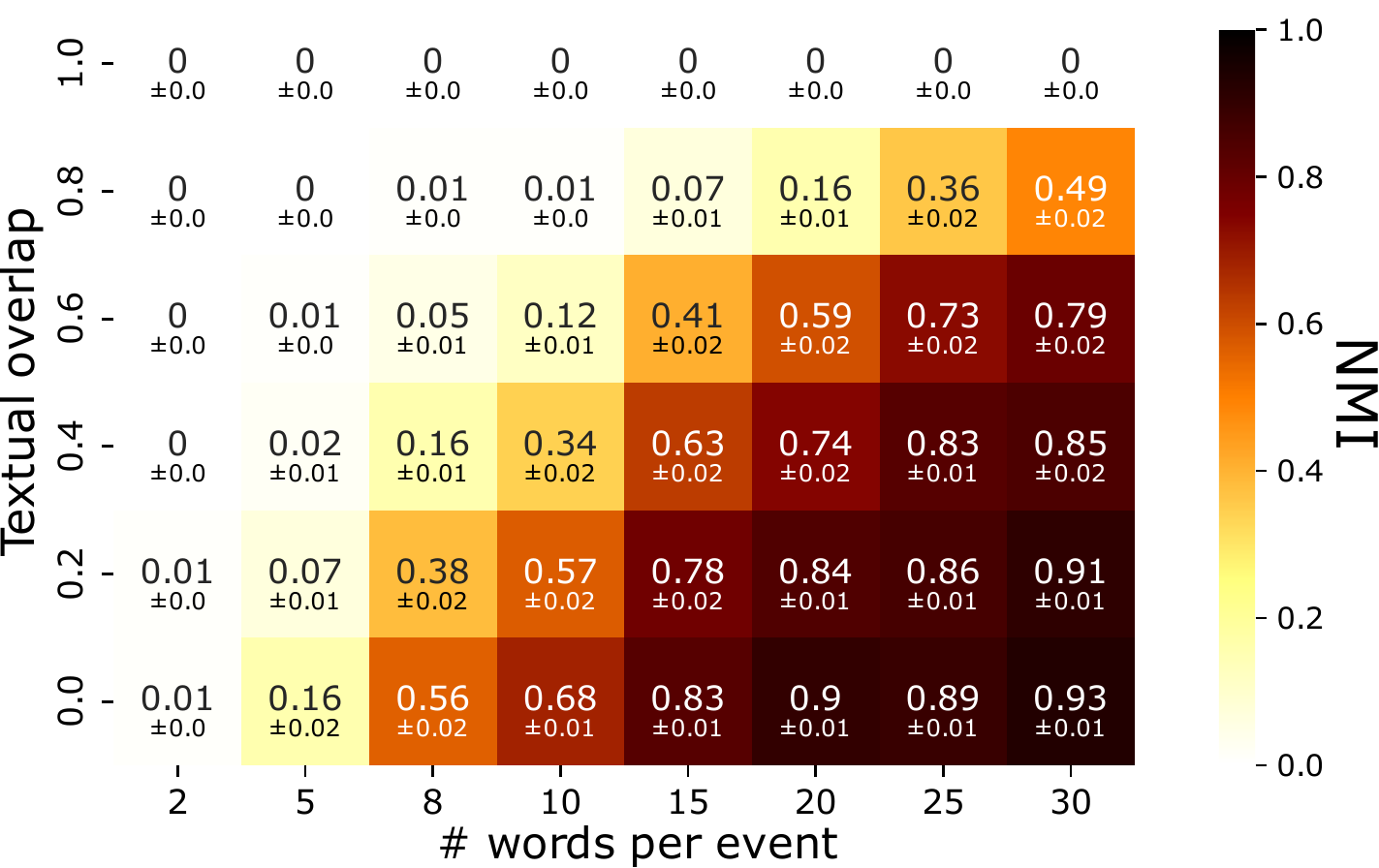}
    \caption{\textbf{How much data to use MPDHP} --- MPDHP performances according to the number of words associated to each event and the overlap between clusters' vocabulary. Overall, MPDHP needs few data. This plot provides a map of how much data must be provided to MPDHP to make it work. For reference, topics overlaps in real world can be estimated around 0.25 \cite{Duran2019OverlapRW} and the number of named entities per Twitter post around 20.}
    \label{fig:XP3}
\end{figure}
}
In this paragraph, we determine how much data should be provided to MPDHP to get satisfying results. In Fig.~\ref{fig:XP2} (right), we plot the performances of MPDHP with respect to the number of words generated by each observation and to the clusters' vocabulary overlap. MPDHP needs a fairly low number of words to yield good results over 5~000 observations. For reference, the overlap between topics can be estimated around 0.25 (\cite{Duran2019OverlapRW}, in Spanish). Similarly, we can estimate an average of $\sim$10-20 named entities per Twitter post (240 characters). These results support the application of MPDHP to model real-world situations.

\subsection{Real-world application on Reddit}
\subsubsection{Data}
We conclude this work with an illustration of MPDHP in a real-world situation. We investigate the interplay between topics on news subreddits, that is how much influence a topic can exert on other ones.
The dataset is collected from the Pushshift Reddit repository \cite{Baumgartner2020PushshiftRedditDataset}. We limit our study to headlines from popular English news subreddits in January 2019: inthenews, neutralnews, news, nottheonion, offbeat, open news, qualitynews, truenews, worldnews. 
From these, we remove posts that have a popularity (difference between upvotes and downvotes) lesser than 20, as they are of lesser influence in the dataset and only add noise to the modeling. 
We remove headlines that contain less than 3 words as they only add noise to the modelling. After curating the dataset in the way described above, we are left with roughly 8,000 news headlines, which makes a total of 65,743 tokens drawn from a vocabulary of size 7,672. Additional characteristics of the dataset are provided as Supplementary Material.

\subsubsection{Parameters}
We run our experiments using a RBF kernel made of Gaussians centred around $\left[ 0, 2, 4, 6, 8 \right]$ hours, with a standard deviation $\sigma$ of 1 hour, $\lambda_0=0.001$, and $r=1$.
We use a Dirichlet-Multinomial language model as in the synthetic experiments with $\theta_0=0.01$. 
As for the SMC algorithm, we set $N_{samples}=100000$. From our observations, the number of coexisting clusters remains around 10 coexisting clusters (roughly 1,000 parameters), allowing sampling each parameter from approximately 100 candidate values.
Each sample parameter is drawn from an identical Beta distribution of concentration parameter $\alpha_0=2$.
We consider 8 particles for the SMC algorithm, similarly to \cite{Du2015DHP}.

\begin{figure}
    \centering
    \includegraphics[width=\textwidth]{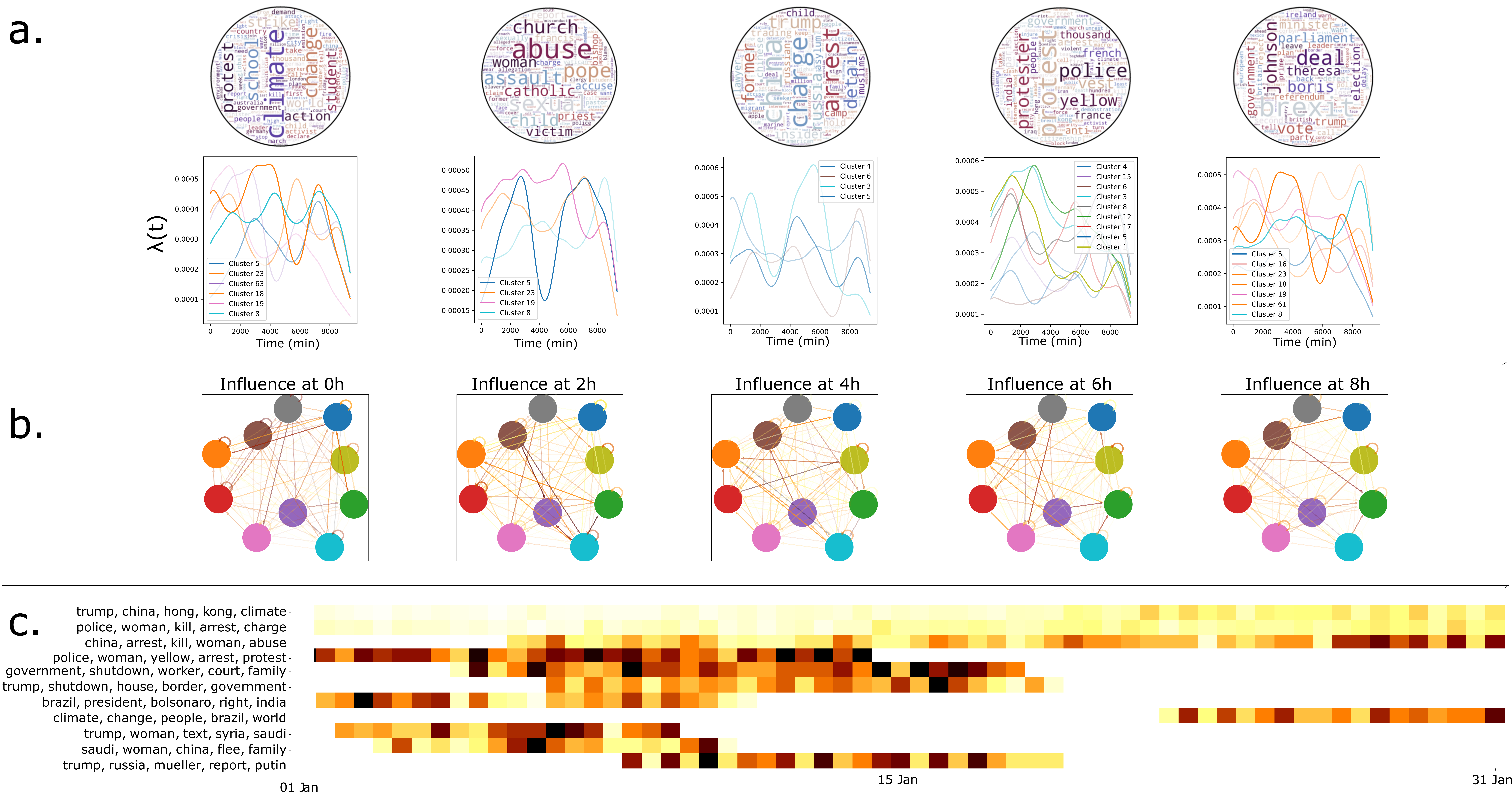}
    \caption{\textbf{Real-world application on Reddit} -- \textbf{a.} Examples of clusters along with their inferred reproduction dynamics. \textbf{b.} Visualisation of interaction patterns at different times as a network; each dot is a cluster, each edge accounts for the value of $\lambda(t)$ at a given time $t \in [ 0;2;4;6;8 ]$ hours. \textbf{c.} Most used clusters represented over real time.}
    \label{fig:appli-reddit}
\end{figure}
\subsubsection{Results}
We present the results of MPDHP on real-world data in Fig.~\ref{fig:appli-reddit}. Fig.\ref{fig:appli-reddit}a. illustrates a typical output from the model. 
The transparency in the representation of $\lambda(t)$ accounts for the number of times such interaction has effectively been observed; transparency of the intensity function $\lambda_{c,c'}(t)$ of $c'$ on $c$ is proportional to $\sum_{t^c} \sum_{t^{c'}<t^{c}} \lambda_{c,c'}(t)$.
We can make two interesting observations from this figure.
Firstly, the interaction strength between clusters seems to fade as time passes (Fig.\ref{fig:appli-reddit}b.). Cluster interactions are more likely to happens within short time ranges.
Secondly, the first two clusters seem to be consistently used across the whole month (Fig.\ref{fig:appli-reddit}c.). When we look at their composition, we notice that the first cluster is made of 75\% of articles from the subreddit r/worldnews, which is +20\% from what one would expect from chance.
Similarly, the second cluster comprises 46\% of r/news articles, which is also roughly +20\% from expected at random. These two clusters therefore significantly account for publications from either of these subreddits, independently from the textual content. Both are general news forums with a large audience; an article that gets posted on other subreddits is likely to also appear on these. Other clusters follow a bursty dynamic, which concurs with \cite{Haralabopoulos2014LifespanInfo}. More details on this experiment can be found in \cite{Poux2022MPDHPPropInterNewsRedditCNA}.

\section{Conclusion}
In this paper, we extended the Powered Dirichlet-Hawkes process so that it can consider multivariate processes, resulting in the Multivariate Powered Dirichlet-Hawkes process (MPDHP). This new process can infer temporal clusters interaction networks from textual data flow. We overcome several optimisation challenges to preserve a time complexity that scales linearly with the dataset.

We showed that MPDHP outperforms existing baselines when clusters interact with each other, and performs at least as well as the PDHP baseline when clusters do not (which PDHP is designed to model). MPDHP can handle cases where textual content is lesser informative better than other baselines. It is robust against tuning of the temporal concentration parameter $\lambda_0$, which allows to handle highly intricate processes. We finally showed that MPDHP performs well with scarce textual data. 
Our results suggest that MPDHP can be applied in a robust way to a broad range of problems, which we illustrate on a real-world application, that provides insights in topical interactions mechanisms between news published on Reddit.

%
%
%
\bibliographystyle{splncs04}
\bibliography{Bibliography}
\end{document}